\documentclass[sigconf,natbib=false, language=english]{acmart}
\usepackage{caption}
\usepackage{subcaption}

\usepackage[datamodel=acmdatamodel, style=acmnumeric, backend=biber]{biblatex}
\AtBeginDocument{%
  \providecommand\BibTeX{{%
    \normalfont B\kern-0.5em{\scshape i\kern-0.25em b}\kern-0.8em\TeX}}}

\setcopyright{rightsretained}
\copyrightyear{2024}
\acmDOI{10.70314/is.2024.scai.4212}

\acmConference[Information Society 2024]{Information Society 2024: 27th international multiconference}{7--11 October 2024}{Ljubljana, Slovenia}
\acmISBN{}
\begin{document}

%\title[Sarcasm Detection]{Sarcasm Detection With Transfer Learning in the Slovene Language}
\title[Sarcasm Detection]{Sarcasm Detection in a Less-Resourced Language}
%% When writing in Slovene, the original Slovenian title should be followed by an English translation.
%\translatedtitle{english}{The Title of the Paper Translated to English}

% \author{Lazar Đoković, Marko Robnik-Šikonja\\ \normalsize
% lazardjokoviclaki02@gmail.com \ \ \ marko.robnik@fri.uni-lj.si}
% \affiliation{%
%   \institution{University of Ljubljana, Faculty of Computer and Information Science}
%   \city{Ljubljana} \country{Slovenia}
% }

\author{Lazar Đoković}
\email{lazardjokoviclaki02@gmail.com}
\affiliation{%
  \institution{University of Ljubljana, Faculty of Computer and Information Science} \city{Ljubljana} \country{Slovenia}
}

\author{Marko Robnik-Šikonja}
\email{marko.robnik@fri.uni-lj.si}
\affiliation{%
  \institution{University of Ljubljana, Faculty of Computer and Information Science}
  \city{Ljubljana} \country{Slovenia}
  }

\renewcommand{\shortauthors}{Đoković and Robnik-Šikonja}


\begin{abstract}

The sarcasm detection task in natural language processing tries to classify whether an utterance is sarcastic or not. It is related to sentiment analysis since it often inverts surface sentiment.   Because sarcastic sentences are highly dependent on context, and they are often accompanied by various non-verbal cues, the task is challenging. Most of related  work  focuses on high-resourced languages like English. To build a sarcasm detection dataset for a less-resourced language, such as Slovenian, we leverage two modern techniques: a machine translation specific medium-size transformer model, and a very large generative language model. We explore the viability of translated datasets and how the size of a pretrained transformer affects its ability to detect sarcasm. We train ensembles of detection models and evaluate models' performance. The results show that larger models generally outperform smaller ones and that ensembling can slightly improve sarcasm detection performance. Our best ensemble approach achieves an $\text{F}_1$-score of 0.765 which is close to annotators' agreement in the source language.

\end{abstract}
% If the language of the document is Slovene, please add an English abstract below and keep the abstract above in Slovenian.
%\begin{translatedabstract}{english}
%	Put the abstract translated into English here.
%\end{translatedabstract}
%% If the text of the paper is in Slovenian, specify the keywords in Slovenian explicitly under the Slovenian abstract above.
%\translatedkeywords{english}{datasets, neural networks, gaze detection, text tagging}


\maketitle

\section{Introduction}

Sentiment analysis is a popular task in natural language processing (NLP), concerned with the extraction of underlying attitudes and opinions, usually categorized as ``positive'', ``negative'', and  ``neutral''.
Detection of sentiment is challenging if the utterances are  ironic or sarcastic. 
{\it Sarcasm} is a form of verbal irony that transforms the surface polarity of an apparently positive or negative utterance/statement into its opposite \cite{sarcasm_analysis}. 
Sarcasm is frequent in our day-to-day communication, especially on social media \cite{new_survey}.
This poses a significant problem for sentiment analysis tools since 
sarcasm polarity switches create ambiguity in meaning.
Sarcasm is highly dependent on its context. For example, the sentence ``{\it I just love hot weather}'' could be interpreted as sarcastic, depending on the situation, e.g., during summer heat waves.

Historical developments of sarcasm detection are surveyed by  \textcite{old_survey}, while recent developments are covered by \textcite{new_survey}.
The problem of automatic sarcasm detection in text is most commonly formulated as a classification task.
Unfortunately, 
sarcasm detection is affected by the lack of large-scale, noise-free datasets. Existing datasets are mostly harvested from  microblogging platforms such as Twitter and Reddit, relying on user annotation via distant supervision through hashtags, such as \textit{\#sarcasm}, \textit{\#sarcastic}, \textit{\#not}, etc. This method is popular since 1) only the author of a post can determine whether it is sarcastic or not, and 2) it allows large-scale dataset creation. However, this method introduces large amounts of noise due to lack of context, user errors, and common misuse on social media platforms.
The sarcasm detection datasets created through manual annotation  tend to be of higher quality but are typically much smaller.
These problems are further compounded for non-English datasets, both manually labeled and automatically collected. Further, as sarcasm strongly relies on its context, using classical machine translation (MT) from English often produces inadequate results. 
This makes sarcasm detection in less-resourced languages, like Slovenian, an even bigger challenge. Therefore, developing reliable sarcasm detection models is of crucial importance for robust sentiment analysis in these languages.

We develop a methodology for sarcasm detection in less-resour\-ced languages and test it on the Slovenian language. We address the problem of missing datasets by comparing state-of-the-art machine translation with large generative models. We explore the viability of such datasets and how the number of parameters affects a model's ability to detect sarcasm. We construct various ensembles of large pretrained language models and evaluate their performance.

The rest of this work is organized as follows.  In Section \ref{approach}, we discuss the proposed approach for detecting sarcasm in a less-resourced language such as Slovenian. We present the creation of a dataset, details of the training methodology and deployed ensemble techniques. We lay out our experimental results and their interpretations in Sections \ref{translation_results} and \ref{model_results}. In Section \ref{conclusion}, we provide conclusions and directions for future work.

\section{Sarcasm Detection Dataset}
\label{approach}

Existing attempts at automatic sarcasm detection have resulted in the creation of datasets in a small number of languages with differing sizes and quality. It is unclear if models trained on these datasets would generalize well to unseen languages \cite{isarcasmEval}. Since no dataset exists for Slovenian, we leverage recent advances in machine translation and large language models (LLMs) to create a dataset for supervised sarcasm detection. We thus apply a translate-train approach when fine-tuning our models.

The prevalence of research done on sarcasm in English means that English datasets are usually larger and of higher quality than their counterparts in other languages. Further, as the translation from (and to) English is usually of better quality, we consider only English datasets. 

Preliminary tests showed poor quality and  poor translation ability of \texttt{Sarcasm on Reddit}\footnote{\href{https://www.kaggle.com/datasets/danofer/sarcasm}{www.kaggle.com/datasets/danofer/sarcasm}} dataset, and \texttt{News Headlines Dataset For Sarcasm Detection}\footnote{\href{https://www.kaggle.com/datasets/rmisra/news-headlines-dataset-for-sarcasm-detection}{www.kaggle.com/datasets/rmisra/news-headlines-dataset-for-sarcasm-detection}}.
Hence, we chose the recent \texttt{iSarcasmEval}\footnote{\href{https://github.com/iabufarha/iSarcasmEval}{github.com/iabufarha/iSarcasmEval}} dataset from the \texttt{SemEval-2022} shared task. We believe that relatively low performance scores obtained in this shared task could be improved with the use of larger LLMs.

\subsection{iSarcasmEval Dataset}
\texttt{iSarcasmEval} is a dataset of both English and Arabic sarcastic and non-sarcastic short-form tweets obtained from Twitter. We use only the English part, which is pre-split into a train and test set. Both sets are unbalanced, the former having 867 sarcastic and 2601 non-sarcastic examples, while the latter has 200 sarcastic and 1200 non-sarcastic examples. 
The authors of the shared task claim that both distant supervision and manual annotation of datasets produce noisy labels in terms of both false positives and false negatives \cite{isarcasmEval}. Thus, they ask Twitter users to directly provide one sarcastic and three non-sarcastic tweets they have posted in the past. These responses are then filtered to ensure their quality.
The produced dataset is not entirely clean since it contains links, emojis, and capitalized text. We chose to leave all of these potential features in the text, as they commonly occur in online conversations and could be indicative of sarcasm.

Let us mention, that an ensemble approach with 15 transformer models and transfer from three external sarcasm datasets proved to be the most accurate modeling technique for English \cite{yuan-etal-2022-stce} achieving an $\text{F}_1$-score of 0.605.

\subsection{Translating iSarcasmEval}

Our preliminary testing using smaller BERT-like classifiers showed that the models learned the distribution of the data and defaulted to the majority classifier ($1200/1400=0.857$ test accuracy). To try to dissuade this, we merged the train and test sets, kept all the sarcastic instances, and randomly sampled an equal number of non-sarcastic examples. This left us with a balanced dataset of 2134 samples.

To enable task specific instructions that would preserve sarcasm, we skipped classical machine translation tools, and tried two alternative translation approaches:
\begin{itemize}
    \item using a medium-sized T5 model trained specifically for neural machine translation,
    \item leveraging a significantly larger model via OpenAI's API.
\end{itemize}

The T5 model uses both the encoder and decoder stacks of the Transformer architecture and is trained within a text-to-text framework. We chose Google's 32-layer T5 model called \texttt{MADLAD400-10B\--MT}\footnote{\href{https://huggingface.co/google/madlad400-10b-mt}{huggingface.co/google/madlad400-10b-mt}}, which has 10.7 billion parameters and is pretrained on the \texttt{MADLAD-400} \cite{madladman} dataset with 250 billion tokens covering 450 languages. Fine-tuning for machine translation was done on a combination of parallel data sources in 157 languages, including Slovenian.

As a generative model, we chose OpenAI's decoder-based \texttt{GPT-4o-2024-05\--13}\footnote{\href{https://platform.openai.com/docs/models/gpt-4o}{platform.openai.com/docs/models/gpt-4o}}. Its true size is not known to the public, but it's speculated that it is significantly smaller than \texttt{GPT-4}, since it is much faster and more efficient. OpenAI claims that it has the best performance across non-English languages of any of their models, thus making it suitable for our task.

When \textit{prompting} generative decoder-based models, it is necessary to craft clear and specific instructions to achieve the best results.  We used few-shot learning \cite{brown2020languagemodelsfewshotlearnersGPT3}, and randomly sampled three training instances, manually translated them, and included them in the following prompt where the double forward slash was used as a delimiter between the query and the expected response.
   
    \textit{You will be provided with a sarcastic/non-sarcastic sentence in English, and your task is to translate it into the Slovenian language. It should keep the original meaning. Examples:} 
    \begin{itemize}
        \item \textit{love getting assignments at 6:25pm on a Friday!! // \\ obožujem, ko mi v petek ob 18:25 pošljejo naloge!!} 
        \item \textit{I still can't believe England won the World Cup // \\ Še vedno ne morem verjeti, da je Anglija zmagala na svetovnem prvenstvu} 
        \item \textit{taking spanish at ut was not my best decision \includegraphics[scale=0.018]{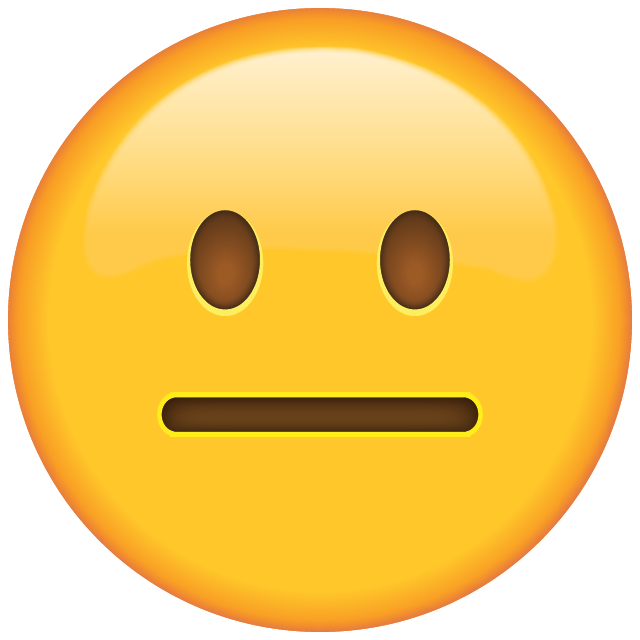} // \\ Učenje španščine na UT ni bila moja najboljša odločitev \includegraphics[scale=0.018]{emojis/neutral-emoji.png}},
    \end{itemize}
We manually assessed the outputs of both transformers in order to determine the best translations for fine-tuning detection models.

\subsection{Translation Results}
\label{translation_results}

During translation, the T5 model sometimes had trouble with examples that had multiple newline characters in a row. It occasionally dropped parts of texts it didn't understand (mostly slang and various types of informal text styles). This shows that a 10B parameter model is not large enough to robustly translate all features of a language such as English into a less-resourced language such as Slovenian. 

On the other hand, the GPT model performed surprisingly well in most instances and it seemed to have a more nuanced understanding of phrases used in online speech. It consistently translated entire texts, keeping the original structure and meaning.
Consequently, we used GPT's translations when training sarcasm detection models.
The translations can be seen in our repository\footnote{\href{https://github.com/GalaxyGHz/Diploma/}{github.com/GalaxyGHz/Diploma}}.

\section{Model Training}

We tested the performance of a wide range of LLMs of different sizes.
Their overview is contained  in Table \ref{params_table}.

\begin{table}[htb]
    \centering
    \caption{Summary of used sarcasm detection models.}
    \begin{tabular}{ l  c } 
        %\hline
        Model & Parameters \\ 
        \hline %\hline
        \texttt{SloBERTa} & 110M  \\
        \texttt{BERT-BASE-MULTILINGUAL-CASED} & 179M  \\ 
        \texttt{XLM-RoBERTa-BASE} & 279M  \\
        \texttt{XLM-RoBERTa-LARGE} & 561M  \\
        \texttt{META-Llama-3.1-8B-INSTRUCT} & 8.03B  \\
        \texttt{META-Llama-3.1-70B-INSTRUCT} & 70.6B  \\
        \texttt{META-Llama-3.1-405B-INSTRUCT} & 406B  \\
        \texttt{GPT-3.5-TURBO-0125} & ?  \\
        \texttt{GPT-4o-2024-05-13} & ?  \\ 
        \hline
    \end{tabular}
    \label{params_table}
    \vspace{-5mm}
\end{table}

\subsection{Encoder Models Under 1B Parameters}
The four smallest models are encoder-based models that embed input text and use a classification head to assign it a class. They required additional fine-tuning to perform sarcasm detection. For these models, we conducted hyperparameter optimization.

We chose the \texttt{SloBERTa} \cite{Ulčar_Robnik-Šikonja_2021-1, Ulčar_Robnik-Šikonja_2021-2} model in order to check whether using a monolingual Slovenian model impacts sarcasm detection performance. We also wanted to compare BERT and RoBERTa-like models, so we used their multilingual variants and fine-tuned them on Slovenian data. 

The models were trained for a maximum of five epochs with the use of early stopping, where the training was halted if the validation loss didn't improve after two epochs.

\subsection{Llama 3.1 Models}

Since the teams that competed in the 2022 shared task on sarcasm mostly used BERT and RoBERTa models, we extend the testing to include significantly larger models. We chose Meta's open-source Llama family of models, more specifically, their newest Llama 3.1 variants. These come in three different sizes, which was perfect for studying the effects of parameter counts on sarcasm detection. We decided to use the ``\texttt{instruct}'' version of all three models since these were fine-tuned to be better at following instructions. 

When prompting LLama and GPT generative models, the following few-shot classification prompt was given, with two positive and two negative examples randomly sampled from our dataset.

\noindent\textit{You will be provided with text in the Slovenian language, and your task is to classify whether it is sarcastic or not. Use ONLY token 0 (not sarcastic) or 1 (sarcastic) as in the examples:} 
\begin{itemize}
    \item \textit{Spanje? Kaj je to... Še nikoli nisem slišal za to? 1} 
    \item \textit{Lepo je biti primerjan z zidom \includegraphics[scale=0.018]{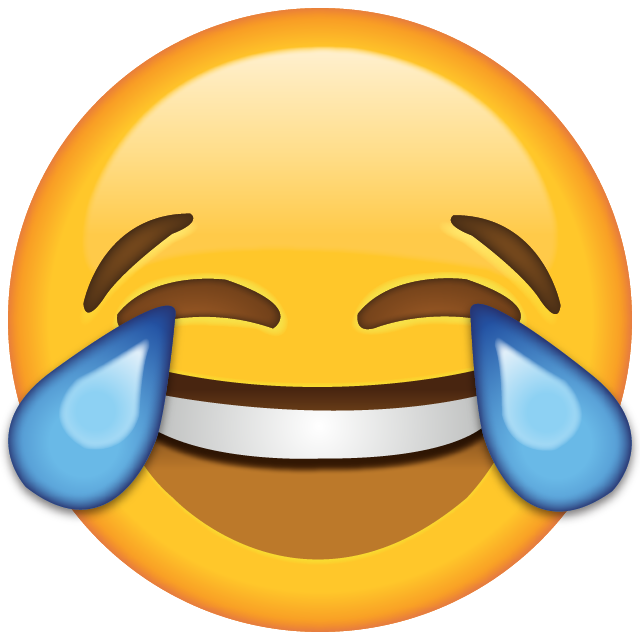} 1} 
    \item \textit{To sploh nima smisla. Nehaj kopati. 0} 
    \item \textit{Dne 12. 10. 21 ob 10:30 je bil nivo reke 0,37 m. 0}.
\end{itemize}

We used full versions of the 8B and 70B parameter models, while the 405B parameter model was loaded in 16-bit precision mode. To minimize the use of resources and costs, we employed LoRA parameter-efficient fine-tuning. We provided the models with training and validation sets and trained them for a maximum of 10 epochs. No hyperparameter optimization was conducted in this case due to time constraints. We used the validation loss to choose the best model, and we used the same early stopping technique as with the smaller models.

\subsection{GPT 3 and 4 Models}

We also tested two models offered on the OpenAI platform, \texttt{GPT-4o-2024-05-13} and \texttt{GPT-3.5-TURBO-0125}. We first used them in few-shot mode and classified all our examples without any fine-tuning. When fine-tuning, the platform's tier system limited us to only the smaller \texttt{GPT-3.5\--TURBO-0125} model. We fine-tuned the model for a maximum of three epochs. In the end, we used the model with the lowest validation loss to classify the examples in the test set.

\subsection{Sarcasm Detection Ensembles}

When constructing ensemble models, we tried two techniques: stacking and voting. In both cases, we used the predicted probability of the sarcastic class from each model as input features.

\subsubsection{Stacking With Regularized Logistic Regression}
Our first ensemble used stacking approach, and logistic regression with Ridge regularization as the meta-level classifier. This choice was motivated primarily by the need for feature selection, as we wanted to identify the most important model predictions and determine which models would be assigned a lower weight. The best models were then used for voting.

\subsubsection{Standard and Mixed Voting}
The second ensembling method was voting. We tried cut-off-based mixed voting inspired by \cite{yuan-etal-2022-stce}. Formally, we used hard voting when the absolute difference between the number of sarcastic and non-sarcastic predictions was greater than $n$, and we used soft voting otherwise. We optimized the value of $n$ based on the ensembles performance on our validation set. 

When $n$ is set to zero, this approach is equivalent to hard voting, and in the case of $n$ being equal to the predictor count, it is equivalent to soft voting. We report both results. Additionally, we compare the results of voting using all trained models with the results obtained by using only the models with large weights in our regularized logistic regression ensemble.

\section{Sarcasm Detection Results}
\label{model_results}

Table \ref{all_results} summarizes all our results. It is roughly sorted by model size, smaller models being on top and larger ones being on bottom. The \texttt{(NFT)} tag indicates that a model was not fine-tuned, while the \texttt{(LoRA)} tag means that a model was trained with LoRA. Results are rounded to three decimal places.

\begin{table}[htb]
    \centering
    \caption{Summary of performance results for all tested models. The best scores are in bold.}
    \resizebox{1.0\columnwidth}{!}{
        \begin{tabular}{ l c c} 
            %\hline
            Model & Accuracy & $\text{F}_1$-score \\ 
            \hline %\hline
            \texttt{SloBERTa} & 0.621 & 0.632  \\
            %\hline
            \texttt{BERT-BASE-MULTILINGUAL-CASED} & 0.499 & 0.666  \\ 
            %\hline
            \texttt{XLM-RoBERTa-BASE} & 0.578 & 0.579  \\
            %\hline
            \texttt{XLM-RoBERTa-LARGE} & 0.550 & 0.597  \\
            \hline
            \texttt{Llama-3.1-8B-INSTRUCT (NFT)} & 0.560 & 0.676  \\
            %\hline
            \texttt{Llama-3.1-8B-INSTRUCT (LoRA)} & 0.569 & 0.682  \\ 
            %\hline
            \texttt{Llama-3.1-70B-INSTRUCT (NFT)} & 0.660 & 0.724  \\
            %\hline
            \texttt{Llama-3.1-70B-INSTRUCT (4-bit-LoRA)} & 0.637 & 0.717  \\
            %\hline
            \texttt{Llama-3.1-405B-INSTRUCT (16-bit-NFT)} & 0.686 & 0.751  \\
            \hline
            
            \texttt{GPT-3.5-TURBO-0125 (NFT)} & 0.564 & 0.679 \\
            %\hline
            \texttt{GPT-3.5-TURBO-0125} & 0.749 & 0.760 \\
            %\hline
            \texttt{GPT-4o-2024-05-13 (NFT)} & 0.686 & 0.746 \\ 
            \hline
            
            \texttt{L2-LOGISTIC-REGRESSION} & \textbf{0.759} & \textbf{0.765} \\ 
            %\hline
            \texttt{L2-LOGISTIC-REGRESSION-NON-COMMERCIAL} & 0.707 & 0.749 \\ 
            %\hline
            \texttt{HARD-VOTING-ALL} & 0.670 & 0.738 \\ 
            %\hline
            \texttt{SOFT-VOTING-ALL} & 0.658 & 0.732 \\ 
            %\hline
            \texttt{HARD-VOTING-BEST-5} & 0.686 & 0.749 \\ 
            %\hline
            \texttt{SOFT-VOTING-BEST-5} & 0.686 & 0.749 \\ 
            \hline
        \end{tabular}
    }
    \label{all_results}
    \vspace{-5mm}
\end{table}

\bigskip

\textbf{Individual Model Performance} \\
Out of all of the used models, only \texttt{BERT-BASE-MULTILINGUAL\--CASED} failed to learn any pattern in our data and defaulted to the dummy classifier.

\texttt{GPT-3.5-TURBO-0125}  sometimes predicts the correct token but then continues to generate additional text, such as 11 and 1\textbackslash n1. This happens with a small quantity of examples in our testing set. We decided to truncate these responses and only kept the first token as the answer.

The Llama models sometimes refused to generate tokens zero or one. We decided to drop these examples altogether. We report the test accuracy and trained the ensemble models without them.

Smaller encoder models performed poorly when compared to some of the larger models. 
Only the \texttt{SloBERTa} model manages to achieve an accuracy above 0.6. Despite being the smallest of the four small models we tested, \texttt{SloBERTa} performed the best. This suggests that the three larger multilingual encoder models may lack sufficient understanding of Slovenian. It also highlights that model size alone does not necessarily correlate with better performance when it comes to sarcasm detection.

The Llama models fared better, achieving accuracies of up to 0.686 with the 405B model being comparable to GPT-4o in performance. They still fell short of the fine-tuned \texttt{GPT-3.5-TURBO-0125} model, which managed to successfully classify about three-quarters of our examples with a $\text{F}_1$-score of 0.76.

Some models had significantly higher $\text{F}_1$-scores and lower accuracies. We show the confusion matrix of one of the models that exhibited the largest difference in Table \ref{tab:confusion_matrix}. These models seem to have a tendency to incorrectly classify non-sarcastic text as sarcastic, leading to a high rate of false positives.

\begin{table}
    \centering
     \caption{Confusion Matrix for non-fine-tuned \texttt{Llama-3.1-405B-INSTRUCT} model.}
    \renewcommand{\arraystretch}{1.5}
    \begin{tabular}{c|cc}
        %\hline
        \textbf{Predicted \textbackslash{} Actual} & \textbf{Positive} & \textbf{Negative} \\ \hline
        \textbf{Positive}  & 202 & 123 \\ %\hline
        \textbf{Negative}  & 11 & 91 \\ \hline
    \end{tabular}
    \label{tab:confusion_matrix}
    \vspace{-5mm}
\end{table}

Our testing also showed that loading the \texttt{Llama-3.1-70B\--INSTRUCT} model in 4-bit mode and fine-tuning it with LoRA does not produce satisfactory results, and it is thus better to conduct full fine-tuning with the smaller Llama model or to use one of OpenAI's models via their fine-tuning API.

\texttt{GPT-3.5-TURBO-0125} performed the best among individual models, so if costs associated with the use of OpenAI's API are acceptable, we recommend its use for sarcasm detection in Slovenian. This shows that very large models can effectively identify sarcasm.
We believe that with better parameter tuning, Llama 8B could be one of the best (and most economical) options for sarcasm detection in Slovenian, provided that the user has sufficient hardware resources.

\textbf{Ensemble Model Performance}\\
We observed that the regularized logistic regression mostly relied on the best-performing models. Its focus on the best model (\texttt{GPT-3.5-TURBO-0125}), however, suggests that there is significant overlap between the various model predictions.

We decided to discard \texttt{BERT-BASE-MULTILINGUAL-CASED} when constructing our voting ensembles since its dummy classification didn't contribute to overall model performance. Both of these two voting classifiers had an odd number of predictors, so there was no need for a tie-breaker mechanism.

Voting proved to be ineffective in our setups, even scoring lower than some of its base models. Hard voting generally outperformed soft voting. We also note that there was no benefit in using mixed voting, at least for the sets of predictors that we obtained as hard voting always had a higher accuracy. This was true for both the classifiers that used all and only five of the base models.

Regularized logistic regression managed to improve on the scores of individual models, raising accuracy by one percent, thus achieving the best performance out of all of the tested approaches. This shows that there is still performance to be gained from ensembles; however, it is still necessary to use commercial models for top performance.

\section{Conclusion}
\label{conclusion}

In this work, we presented the task of sarcasm detection in the less-resourced Slovenian language. Our code and results are freely available\footnote{\href{https://github.com/GalaxyGHz/Diploma/}{github.com/GalaxyGHz/Diploma}}.

We tackled the missing dataset problem by using two LLMs to perform neural translation of an English dataset into Slovenian. The translations generated by \texttt{GPT-4o-2024-05-13} outpaced those generated by a large T5 model specifically trained for neural machine translation in terms of quality.

We used this data to train a plethora of Transformer-based models in various settings. We found that fine-tuning \texttt{GPT-3.5\--TURBO-0125} via OpenAI's API results in the highest individual Slovenian sarcasm detection power, but we also note that a possible alternative could be local fine-tuning of the \texttt{Llama-3.1-8B\--INSTRUCT} model. Our testing shows that using aggressive quantization combined with LoRA results in significant performance degradation.

We also constructed ensemble models based on voting and stacking methods. Observations showed that voting didn't result in any performance improvements. On the other hand, stacking with the use of a regularized logistic regression managed to improve on the performance of its base models.

Additional work needs to be done in dataset construction.
Sarcastic examples could be extended with context or labels of the types of sarcasm they represent. This might help guide models towards better understanding of sarcasm.
Future work could also explore incorporating heterogeneous models into ensembles or the creation of Mixture of Experts (MoE) ensembles, whose baseline models would focus on different aspects of sarcasm.

% \section{Appendices}

% If your work needs an appendix, add it before the
% ``\verb|\end{document}|'' command at the conclusion of your source
% document.

% Start the appendix with the ``\verb|appendix|'' command:
% \begin{verbatim}
%   \appendix
% \end{verbatim}
% and note that in the appendix, sections are lettered, not
% numbered. This document has two appendices, demonstrating the section
% and subsection identification method.

% %%
% %% The acknowledgments section is defined using the "acks" environment
% %% (and NOT an unnumbered section). This ensures the proper
% %% identification of the section in the article metadata, and the
% %% consistent spelling of the heading.
\begin{acks}
This research was supported by the Slovenian Research and Innovation Agency (ARIS) core research programme P6-0411 and projects J7-3159, CRP V5-2297,  L2-50070, and PoVeJMo.
\end{acks}

\printbibliography

% %%
% %% If your work has an appendix, this is the place to put it.
% \appendix









\end{document}